\documentclass[10pt,twocolumn,letterpaper]{article}

\usepackage{cvpr}              % To produce the CAMERA-READY version
\usepackage{amsmath,amsfonts}
\usepackage{algorithmic}
\usepackage{algorithm}
\usepackage{array}
\usepackage{textcomp}
\usepackage{stfloats}
\usepackage{url}
\usepackage{verbatim}
\usepackage{graphicx} 
\usepackage{natbib}
\usepackage{graphicx}
\usepackage{bm}
\usepackage{amsmath}
\usepackage{amssymb}
\usepackage{epsfig}
\usepackage{graphicx}

\usepackage{amsmath}
\usepackage{amssymb}
\usepackage{mathtools} 
\usepackage{bm}
\usepackage{algorithm}
\usepackage{algorithmic}
\usepackage{multirow,booktabs} 
\usepackage{amssymb}
\usepackage{tabu}
\usepackage{array}
\usepackage{algorithm}
\usepackage{algorithmic}
\usepackage{times}
\usepackage{xcolor,colortbl}
\usepackage{epsfig}
\usepackage{graphicx}
\usepackage{amsmath}
\usepackage{multirow}
\usepackage{amssymb}
\usepackage{booktabs}       % professional-quality tables
\usepackage{amsfonts}       % blackboard math symbols
\usepackage{nicefrac}       % compact symbols for 1/2, etc.
\usepackage{microtype}      % microtypography
\usepackage{amsmath}
\usepackage{wrapfig}
\usepackage{amssymb}
\usepackage{color}
\usepackage{marvosym}

\usepackage{marvosym}

\newtheorem{theorem}{Theorem}

\newtheorem{remark}{Remark}

\definecolor{iccvblue}{rgb}{0.21,0.49,0.74}

\definecolor{darkgreen}{rgb}{0.0, 0.2, 0.13}

\definecolor{exp_table_blue}{HTML}{DAECED}

\definecolor{cvprblue}{rgb}{0.21,0.49,0.74}
\usepackage[pagebackref,breaklinks,colorlinks,allcolors=cvprblue]{hyperref}

\def\paperID{2081} % * Enter the Paper ID here
\def\confName{CVPR}
\def\confYear{2026}

\title{Dataset Distillation by Influence Matching}

\author{Haoru Tan$^{1,\dagger}$~~~~~~~~~~~~~~~~~~Wang Wang$^{1,\dagger}$~~~~~~~~~~~~~~~~~~Sitong Wu$^{2}$~~~~~~~~~~~~~~~~~~Xiuzhe Wu$^{3}$\\
\\
Yang-Tian Sun$^{1}$~~~~~~~~~Chirui Chang$^{1}$~~~~~~~~~Shaofeng Zhang~~~~~~~~~Xiaojuan Qi$^{1, \textrm{\Letter}}$\\
\\
$^{1}$HKU~~~~~~~~~~~~~~~~~~~~~~~~~~$^{2}$CUHK~~~~~~~~~~~~~~~~~~~~~~~~~~$^{3}$Stanford\\ 
\\
{\tt\small hrtan@eee.hku.hk} 
}
\begin{document}
\maketitle

\newcommand\blfootnote[1]{%
  \begingroup
  \renewcommand\thefootnote{}\footnote{#1}%
  \addtocounter{footnote}{-1}%
  \endgroup
}

\blfootnote{$^{\dagger}$ Equal first author.}

\begin{abstract}

We revisit dataset distillation from an outcome-centric perspective. Rather than aligning process surrogates (per-step gradients or training trajectories), Influence Matching (Inf-Match) aligns the final outcome of training: it learns a compact synthetic set whose effect on the converged parameters matches that of the full dataset. Concretely, we introduce a fully differentiable, sample-level influence estimator that quantifies parameter shifts from adding or removing data, without time-consuming inverse-Hessian products or convexity assumptions. The estimator runs in linear time by unrolling the optimization dynamics and applying a first-order Taylor approximation. We then learn the synthetic set by minimizing the mismatch between its influence and that of the real dataset, yielding outcome alignment rather than heuristic process imitation. Inf-Match delivers the best accuracy across standard classification benchmarks. For instance, on Tiny-ImageNet (IPC=10), Inf-Match attains 31.5\%, a +4.7\% improvement over NCFM. Beyond classification, Inf-Match scales to vision-language distillation on Flickr30K, outperforming strong process-matching baselines. For instance, with 200 to 1000 synthetic samples, our method achieved a leading impressive average on image/text retrieval tasks, higher than NCFM by 2.5\%. The code will be released via \textcolor{teal}{https://github.com/hrtan/infmatch}.

\end{abstract}

\section{Introduction}
\label{sec:introduction} 

With the rapid expansion of visual data across domains such as autonomous driving, surveillance, and web-scale imagery, the size of modern datasets has grown to tens or even hundreds of millions of samples. While large-scale data fuels the success of deep vision models, it also brings prohibitive costs in storage, transmission, and training. These challenges have motivated dataset distillation, the task of synthesizing a small but highly informative dataset that encapsulates the knowledge of a much larger one \cite{wang2018dataset,li2022awesome,zhaodm,zhaodsa}. By learning a handful of representative samples that preserve the training dynamics or final performance of the full dataset, dataset distillation enables efficient data sharing \cite{zhaodm,zhaogm}, rapid model adaptation, privacy-preserved learning \cite{dong2022privacy}, and continual or federated learning \cite{yang2023efficient,gu2023summarizing} under strict resource budgets.

% \begin{figure}[tp] 
% \centering  
% \includegraphics[width=0.999936\linewidth]{fig/cvpr_radar.pdf}
% \vspace{-0.8cm}
% \caption{\label{fig:intro}
% The final performance comparison between ours and the two other baselines, namely DATM \cite{guo2023towards} and NCFM \cite{wang2025datasetdistillationneuralcharacteristic}, on image classification datasets, including CIFAR \cite{CIFAR} and Tiny-ImageNet \cite{tiny}, and the multimodal vision-language dataset Flickr30K \cite{flickr30k}. The performance on Flickr30K is the average result of the Image-to-Text and Text-to-Image retrieval experiments under different NoP (number of pictures) settings. For detailed results, please refer to Table \ref{tb:vl}. } 
% \end{figure}

\begin{figure*}[tp] 
\centering  
\includegraphics[width=0.999998199936\linewidth]{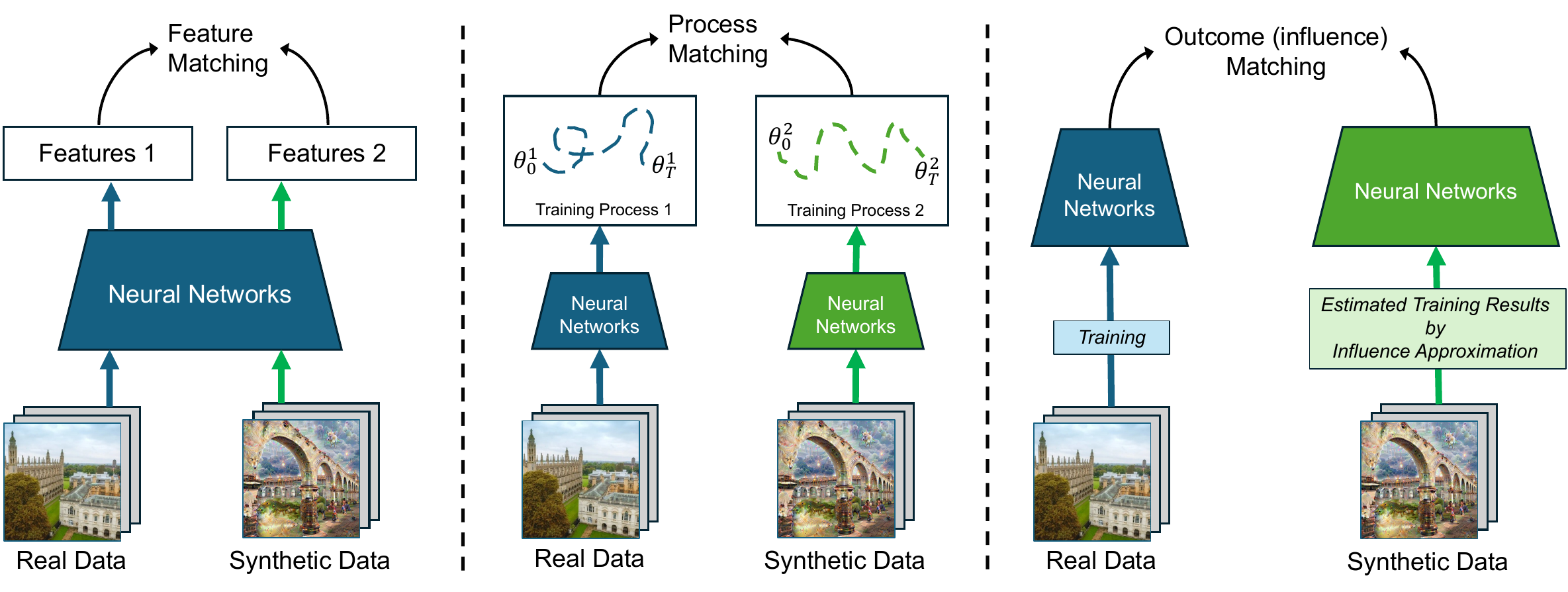} 
\caption{\label{fig:pipeline} This figure outlines three core paradigms for data comparison or generation: (a) Feature matching, where a feature extractor is trained on real data, and synthetic data is generated to match the extracted features \cite{zhaodm,zhaodsa,wang2025datasetdistillationneuralcharacteristic}; (b) (Optimization) Process matching, which seeks to align the optimization paths \cite{cazenavette2022dataset,guo2023towards} or the gradients \cite{zhaogm} of models trained on synthetic and real data; and (c) Our outcome matching pipeline, which focuses on matching the final trained models (outcome) resulting from both synthetic and real data. Crucially, the model trained on synthetic data in our methodology is not obtained through actual retraining; rather, its parameters are effectively estimated by leveraging our novel influence estimator, see Remark \ref{remark} for details. 
} 
\end{figure*}

Early efforts tackle this problem by approximating the original bilevel optimization of dataset distillation (See Eq. \eqref{eq: goal dd}) with more tractable proxies, like feature matching \cite{zhaodsa,wang2022cafe,zhaodm} and (optimization) process matching \citep{zhaogm,cazenavette2022dataset}. The latter has received more attention due to its generally superior performance. Typical process matching approaches include Gradient Matching (GM) \cite{zhaogm}, which aligns per-step gradients between real and synthetic data, and Trajectory Matching (MTT) \cite{cazenavette2022dataset}, which enforces consistency between training trajectories. These methods have achieved encouraging progress, yet they rely on heuristic surrogates that only mimic intermediate training behaviors. Consequently, the synthetic data may perfectly reproduce the training process of the real data without guaranteeing comparable performance or generalization. 

Despite recent advancements such as improved feature distribution matching \cite{wang2025datasetdistillationneuralcharacteristic} or difficulty-aware trajectory alignment \cite{guo2023towards}, a fundamental limitation persists: these alignment proxies do not imply outcome alignment. The ultimate goal of dataset distillation is not to match training steps but to reproduce the influence that the full dataset exerts on the learned model. Bridging this \textit{optimization gap} demands a principled way to quantify how individual samples or subsets influence the final trained model. However, existing influence estimators are computationally prohibitive \cite{IF,Studying_large_language,kwon2023datainf,group_rffect_analysis,group_rffect_improve,Arnoldi}, requiring inverse-Hessian computations and assuming convex losses that fail for deep neural networks. As a result, a key open question remains: can we directly distill data by matching its outcome influence on the final model, rather than its intermediate dynamics?

We address this question by introducing \textbf{Influence Matching (Inf-Match)}, a new dataset distillation method that shifts the focus from process alignment to \textit{outcome alignment}. The core of our framework is a \textit{differentiable sample influence estimator} that quantifies how a data sample or a group of samples contributes to the final optimized model parameters without any convexity assumptions or inverse-Hessian computation. without convexity assumptions and without inverse–Hessian computations. We derive this estimator by unrolling the optimization dynamics and applying a first-order Taylor approximation, yielding \emph{linear-time} complexity and high-fidelity estimates that are practical for scalable real-world applications. 
Then, \textbf{Inf-Match} optimizes the synthetic dataset so that its influence on the model matches the influence of the original dataset, effectively ensuring that training on the synthetic data yields the same final model as training on the full data. This formulation markedly narrows the optimization gap in dataset distillation and establishes a direct path toward outcome-aligned distillation.

Extensive experiments validate the effectiveness and generality of our approach. On CIFAR-10, CIFAR-100, and Tiny-ImageNet, \textbf{Inf-Match} consistently achieves the best performance, surpassing strong baselines such as DATM \cite{guo2023towards} and NCFM \cite{wang2025datasetdistillationneuralcharacteristic} across all image-per-class (IPC) settings, see Figure \ref{tb:cls}.  
For instance, on Tiny-ImageNet (IPC=10), \textbf{Inf-Match} attains 31.5\%, a +4.7\% improvement over NCFM. 
It further demonstrates strong scalability to vision-language datasets such as Flickr30K. 
For instance, with 200 synthetic samples, our method achieved an impressive score on image-to-text retrieval tasks that is higher than the next best, DATM \cite{guo2023towards}, at 1.3\%. Moreover, with 200 to 1000 synthetic samples, our method achieved a leading impressive average on image/text retrieval tasks, higher than NCFM by 2.5\%, see Figure \ref{tb:vl}.

\section{Related Works}

\noindent \textbf{Dataset Distillation} Given the burdensome nature of the original problem in Eq. \eqref{eq: goal dd}, one has to explore proxy tasks, such as {feature matching} \cite{zhaodm,zhao2023improved,sajedi2023datadam,zhang2024m3d} and {(optimization) process matching} \cite{zhaodsa,feng2023embarrassingly,du2024sequential,cazenavette2022dataset,du2023minimizing,cui2023scaling,guo2023towards}. In the following, we mainly review these works. 

\textit{(1). Feature matching.}   
There is one line of work \cite{zhaodm,zhao2023improved,sajedi2023datadam,zhang2024m3d} that tries to match the latent feature space directly. 
Distribution matching (DM) \cite{zhaodm} proposed to match the synthetic and target data from the distribution perspective for dataset distillation. 
CAFE \cite{wang2022cafe} improved the distribution matching from several aspects: (1) using multiple-layer features other than only the last-layer features for matching, (2) proposing the discrimination loss to enlarge the class distinction of synthetic data. 
IDM \cite{zhao2023improved} adds a classification loss as regularization to mitigate less classified synthetic data caused by the first-order moment mean matching. 
Datadam \cite{sajedi2023datadam} proposed to learn synthetic images by matching the spatial attention maps of real and synthetic data generated by different layers within a family of randomly initialized neural networks. 
M3D \cite{zhang2024m3d} proposed to minimize the maximum mean discrepancy (MMD) between the real and the synthetic data. 

\textit{(2). Process matching.}  
Another group of methods \cite{zhaogm,zhaodsa,du2024sequential,cazenavette2022dataset,du2023minimizing,cui2023scaling,guo2023towards} constructs the surrogate problem of matching the intermediate training state contributed by the synthetic data and the real data, respectively. 
Among them, the most representative schemes are matching gradient \cite{zhaogm} in training and matching trajectory \cite{cazenavette2022dataset} in training. 
DSA \cite{zhaodsa} proposed incorporating the gradient matching framework with a differentiable augmentation scheme to synthesize more informative synthetic images and for better performance when training networks with augmentations. 
SeqMatch \cite{du2024sequential} addresses the issue of failing to condense high-level features in dataset distillation. It divides synthetic data into multiple subsets, sequentially optimizing them to promote the effective distillation of high-level features learned in later epochs. 
MTT \cite{cazenavette2022dataset} proposes a new formulation that optimizes our distilled data to guide networks to a similar state as those trained on real data across many training steps. DATM \cite{guo2023towards} also distills easy/difficult information into the trajectory matching framework, achieving further performance improvements.

\textit{(2). Others.}  Beyond the above, some recent studies \cite{yin2024squeeze,sun2023diversity} have integrated considerations of data diversity and the authenticity of synthesized data into their framework designs. Additionally, research \cite{cazenavette2023generalizing,zhang2023dataset,wang2023dim} has investigated the applicability of generative models in dataset distillation. Furthermore, both meta-gradient-based methods \cite{wang2018dataset,sucholutsky2021soft,deng2022remember} and kernel-based methods \cite{nguyen2020dataset,zhou2022dataset,loo2022efficient} have been explored in dataset distillation. 
It is worth noting that Kernel-based methods \cite{nguyen2020dataset,zhou2022dataset,loo2022efficient} can theoretically estimate the results of inner-loop training directly, avoiding the need for inner-loop training. However, a significant challenge is that by approximating the learning process with a linear kernel, the kernel may overlook the complex training dynamics \cite{atanasov2021neural}. This, in turn, results in a decline in overall performance. 

\vspace{0.1in}\noindent\textbf{Influence Estimation.} Influence estimation \cite{survey,cook1986assessment,IF,shapley1953value,kernel_shapley,tan2023data,tan2025understandingdatainfluencedifferential,tan2025understanding} focuses on linking the training data to the performance of a model post-training. A common approach to assessing the impact of a specific data point is through leave-one-out (LOO) retraining. This technique involves training the model on the dataset while omitting certain samples, then evaluating the changes in performance compared to the model trained with the complete dataset. 
Instead of relying on full retraining, Koh et al. \cite{IF} suggest an alternative method that estimates the effects on the model from slight modifications in the weights of the training data. Their approach utilizes an estimator derived from the product of the inverse Hessian and the gradient, allowing for an efficient approximation of the influence exerted by individual samples. 
In subsequent research, various initiatives have sought to enhance this method along different avenues. To improve scalability, several approaches have been introduced \cite{Studying_large_language,Arnoldi,kwon2023datainf}. 
Regarding the estimation precision, some studies have focused on analyzing \cite{group_rffect_analysis} and refining \cite{group_rffect_improve} the influence function to better assess group impacts. 

However, these works have some significant shortcomings. Firstly, they depend on the rather stringent assumption that the loss function is convex to parameters, a condition that is frequently not met  \cite{choromanska2015loss,dauphin2014identifying}. Secondly, scaling these methods to accommodate large models and extensive datasets poses challenges, primarily due to the computational burden associated with the inverse-Hessian-gradient product. These factors significantly restrict their applicability.

\section{Preliminaries} 
Given a real dataset $\mathcal{D}$, a deep network with parameters $\boldsymbol{\theta}$, and a task loss $\mathcal{L}(\cdot,\cdot)$, \emph{dataset distillation} seeks a compact synthetic set $\mathcal{S}$ such that training on $\mathcal{S}$ yields a model that performs comparably to one trained on $\mathcal{D}$. This is naturally posed as a bilevel program~\cite{wang2018dataset,zhaogm}: 
\begin{equation}
\underbrace{\mathcal{S}^* =\arg\min_{\mathcal{S}} \mathcal{L}(\mathcal{D}, \bm{\theta}^{*}_{\mathcal{S}})}_{\small\text{Outer-Level: Data Optimization}} ~~~ \text{s.t.}~~~
\underbrace{\bm{\theta}^{*}_{\mathcal{S}} = \arg\min_{\bm{\theta}} \mathcal{L}(\mathcal{S}, \bm{\theta})}_{\small\text{Inner-Level: Network Optimization}}, 
\label{eq: goal dd}
\end{equation} 
where the inner problem trains the network on $\mathcal{S}$ to obtain $\boldsymbol{\theta}^{*}_{\mathcal{S}}$, and the outer problem updates $\mathcal{S}$ so that the resulting model minimizes loss on $\mathcal{D}$. In effect, the distilled set $\mathcal{S}$ is learned to induce a final model whose performance closely mirrors that of training on the full dataset $\mathcal{D}$.

Dataset distillation is, by nature, a bi-level optimization problem, which is notoriously hard to solve in practice. Consequently, many methods resort to proxy objectives \cite{zhaodm,zhaogm,guo2023towards}, such as trajectory matching \cite{cazenavette2022dataset} or gradient alignment \cite{zhaogm}. These proxies rest on a heuristic assumption: if the synthetic data can mimic intermediate training signals (e.g., gradients, parameter states), it will yield the same outcome as the full dataset. This assumption is fragile. It induces an optimization gap in which synthetic data can excel at the proxy objective, yet still underperform on downstream accuracy and generalization. In effect, there is a fundamental trade-off between computational tractability (via process alignment) and fidelity to the true objective (matching the final, outcome-level behavior of the model).

\section{Method} 

\label{sec: method}

Here, we present the formulation of our proposed dataset distillation framework, \textbf{Influence Matching (Inf-Match)}. We begin by defining the influence of individual samples or subsets of data on the final trained model and then introduce our efficient estimator.

\subsection{Data Influence} 

Given a model with parameters $\bm{\theta}$ to be trained, and an original training dataset $\mathcal{D}$, the \textbf{removal} influence of a single sample or a group of samples $\mathcal{Z} \subset \mathcal{D}$, denoted as $\mathcal{I}_{-\mathcal{Z}}$, is defined by the difference in the model's final parameters when trained without and with $\mathcal{Z}$: 
$$
\textbf{Removal-Influence:}~~~\mathcal{I}_{-\mathcal{Z}} = \bm{\theta}^*_{\mathcal{D}-\mathcal{Z}} -\bm{\theta}^*_{\mathcal{D}},
$$
where $\bm{\theta}^*_{\mathcal{D}}$ represents the optimal parameters of the model trained on the full dataset $\mathcal{D}$, and $\bm{\theta}^*_{\mathcal{D}-\mathcal{Z}}$ represents the optimal parameters of the model trained on the dataset with the set $\mathcal{Z}$ removed. Intuitively, $\mathcal{I}_{-\mathcal{Z}}$ quantifies the exact change in the model's final state that is attributable to the presence of the data $\mathcal{Z}$ during training.  
Analogously, we define the influence resulting from the \textbf{addition} of a set of external samples $\mathcal{Z} \not\subset \mathcal{D}$ (which originally do not belong to the training set $\mathcal{D}$) as: 
$$\textbf{Addition-Influence:}~~~
\mathcal{I}_{+\mathcal{Z}} = \bm{\theta}^*_{\mathcal{D}+\mathcal{Z}} -\bm{\theta}^*_{\mathcal{D}},
$$
where $\bm{\theta}^*_{\mathcal{D}+\mathcal{Z}}$ represents the optimal parameters of the model trained on the combined dataset $\mathcal{D} \cup \mathcal{Z}$. This quantity $\mathcal{I}_{+\mathcal{Z}}$ captures the exact shift in the final model parameters caused by introducing the new data $\mathcal{Z}$.

\paragraph{Data Influence Estimator.} Existing influence estimators~\cite{IF,kwon2023datainf,Studying_large_language,group_rffect_analysis,Arnoldi} suffer from two major limitations: (i) they rely on the convexity of the loss landscape, which rarely holds for deep networks, and (ii) they require computing inverse–Hessian products, leading to high computational overhead~\cite{IF,kwon2023datainf}.
To overcome these issues, we develop an efficient, fully differentiable estimator by unrolling the optimization dynamics and applying a first-order Taylor approximation. It achieves \emph{linear-time} complexity, avoids any inverse–Hessian computation by Theorem \ref{theorem: 1}, and is supported by a provable (tight) upper bound on the estimation error, see Theorem \eqref{theorem: 2} for more details.

\begin{theorem} 
\label{theorem: 1} \textbf{(Influence estimator)}
{Let $\{(\bm{\theta}^t_{\mathcal{D}}, \eta_t) |_{t=1}^T \}$ denote a series of parameters and learning rates used during the model training on $\mathcal{D}$ with the SGD optimizer. 
Let $H$ denote the Hessian and $G$ indicate the gradient, respectively. 
Specifically, we have $H^t_{\mathcal{D}} = \nabla^2_\theta \mathcal{L}(\mathcal{D}, \theta^t_\mathcal{D})$ and $H^t_{\mathcal{Z}} = \nabla^2_\theta \mathcal{L}(\mathcal{Z}, \theta^t_\mathcal{D})$, moreover, $G^t_{\mathcal{D}} = \nabla_\theta \mathcal{L}(\mathcal{D}, \theta^t_\mathcal{D})$ and $G^t_{\mathcal{Z}} = \nabla_\theta \mathcal{L}(\mathcal{Z}, \theta^t_\mathcal{D})$. 
For $\mathcal{Z} \subset \mathcal{D}$, the removal influence and the addition influence could be efficiently estimated via:} 
\begin{align}
\mathcal{I}_{-\mathcal{Z}} &\approx - \sum_{t} \frac{\eta_t  \sum_{k \geq t} \eta_k}{|\mathcal{D}|}  \Big( H^t_\mathcal{D} G^t_{\mathcal{Z}}  +  H_{\mathcal{Z}}^t G_\mathcal{D}^t\Big),\label{eq:proposition removal}\\
\mathcal{I}_{+\mathcal{Z}} &\approx~~ \sum_{t} \frac{\eta_t  \sum_{k \geq t} \eta_k}{|\mathcal{D}|}  \Big( H^t_\mathcal{D} G^t_{\mathcal{Z}}  +  H_{\mathcal{Z}}^t G_\mathcal{D}^t\Big). 
\label{eq:proposition addition}
\end{align} 
\end{theorem} 
The estimator leverages parameter checkpoints along the SGD training trajectory $t$ on $\mathcal{D}$. Although second-order in form, the Hessian-gradient product can be efficiently approximated via the established technique~\cite{fast_hessian}: 
\begin{equation} 
H G \approx \lim_{\epsilon \rightarrow 0} \frac{\Big(\nabla_\theta \mathcal{L}(\theta + \epsilon G) - \nabla_\theta \mathcal{L}(\theta)\Big)}{\epsilon} ,
\label{eq: hessian}
\end{equation}
where the computational complexity is $\mathcal{O}(p)$, with $p$ denoting the number of parameters. This can be executed efficiently using popular deep learning frameworks \cite{pytorch}. 
%\vspace{0.1in}\noindent \textit{Approximation error analysis.}  
We further establish a theoretical upper bound on the approximation error between our estimator and the exact influence obtained through full retraining. This bound guarantees robustness even under worst-case conditions and demonstrates strong practical reliability compared with prior estimators~\cite{hara2019data,IF,TracIN}.

\begin{theorem}
\label{theorem: 2} \textbf{(Error bound)}
{Let $T$ denote the maximum iteration.  
By supposing the gradient of the loss is $\ell$-Lipschitz continuous and the gradient norm of the network parameter is upper-bounded by $g$, and denote the maximum learning rate by $\eta_\text{max}$. 
The approximation error between the estimator ($\tilde{\mathcal{I}}$) and the exact one (denoted by ${\mathcal{I}}$) is bounded by:  }
\begin{equation} 
|{\tilde{\mathcal{I}}} - \mathcal{I}| \leq 2 T^3\ell ( T + 1) \eta_\text{max} g + \frac{|\mathcal{Z}|}{|\mathcal{D}|}T^2g. 
\end{equation} 
\end{theorem}

From this theorem, several observations follow:
(i) The Lipschitz constant $\ell$ and gradient norm $g$ jointly control the estimation error-- models with more stable (less rapidly changing) gradients yield tighter bounds. 
(ii) The bound scales polynomially with training steps $T$, improving upon earlier estimators~\cite{hara2019data,schioppa2024theoretical} whose error grows exponentially.
(iii) Although the bound represents a worst-case scenario, empirical results indicate that the estimated influence correlates closely with the exact influence across realistic settings. This reliability stems from the incremental nature of SGD updates, which stabilizes training dynamics and maintains the estimator’s fidelity even for long optimization horizons.

\begin{algorithm*}[tp]
\caption{Dataset distillation by influence matching (\textbf{Inf-Match})}
\label{algo: Pipeline} 
\begin{algorithmic}[1]
%\SetKwInOut{KIN}{Input}
%\SetKwInOut{KOUT}{Output}
%\SetKwInOut{INITIAL}{Initialize}
\STATE \textbf{Input:} 
$\mathcal{D}$: original dataset; $\mathcal{T} = \{(\theta^t_\mathcal{D}, \eta_t) |_{t=1}^T \}$: a training trajectory of a network trained on $\mathcal{D}$; and
$t_m$: the number of sampled time-steps. \\
    \vspace{0.05cm}
\STATE \textbf{Initialization:} Initialize the synthetic set $\mathcal{S}=\{(x_i,\hat{y}_i)|x_i\in\mathcal{D}$\}, where $\hat{y}_i=f(x_{i};\theta^T_\mathcal{D})$ is the soft-label from the learned model $\theta^T$. \\ 
    \vspace{0.05cm}
\FOR{ $t$  {from} {\rm 0} to {\rm max\_iteration}} 
    \vspace{0.05cm}
\STATE     Randomly sample a minibatch $B_\mathcal{S} \subset \mathcal{S}$, $B_\mathcal{D} \subset \mathcal{D}$. \\ 
%\xjqi{how do you train the model, what is the training set}
    \vspace{0.05cm}
\STATE     Sample checkpoints $\{\theta^{t_1}_{\mathcal{D}},...,\theta^{t_m}_{\mathcal{D}}\}$ from \( m \) time steps from  training trajectory $\mathcal{T}$.\\
    \vspace{0.05cm}
\STATE     Compute the loss via Eq.\eqref{eq: new problem} on sampled checkpoints and  $B_\mathcal{S}$ and $B_\mathcal{D}$.\\
    \vspace{0.05cm}
\STATE     Update $\mathcal{S}$ (for both images and soft labels) by minimizing Eq.\eqref{eq: new problem} with gradient descent.\\
    \vspace{0.05cm}
\ENDFOR
\STATE \textbf{Output:} The learned synthetic set $\mathcal{S}$. 
    \vspace{0.05cm}
\end{algorithmic}
\end{algorithm*}

\subsection{Dataset Distillation via Influence Matching}   
\label{sec:Objective Formulation}

Based on the influence estimator above, we formulate dataset distillation as learning a synthetic dataset $\mathcal{S}$ that can substitute the full real dataset $\mathcal{D}$ in terms of its overall influence on the model parameters. 
Intuitively, the influence of adding the synthetic data $\mathcal{S}$ should ideally offset the influence of removing the entire real dataset $\mathcal{D}$. 
Formally, this can be expressed as minimizing the total parameter shift:
\begin{equation}
\mathcal{S}^*=\arg\min_{\mathcal{S}} \Big\|\mathcal{I}_{-\mathcal{D}} + \mathcal{I}_{+\mathcal{S}}\Big\|,
\label{eq: new object}
\end{equation}
where $|\cdot|$ denotes a vector norm (typically $L_2$), measuring the magnitude of the residual parameter difference.
By minimizing this quantity, we ensure that the model trained on $\mathcal{S}$ yields parameters closely aligned with those trained on $\mathcal{D}$-- achieving outcome alignment without retraining.

\begin{remark} \label{remark}
The objective function defined in $\text{Eq.}\eqref{eq: new object}$ establishes a novel, principled optimization goal: it aims to learn a synthetic dataset $\mathcal{S}$ such that the resulting model trained on $\mathcal{S}$ achieves an outcome (parameter update) that aligns with the outcome achieved by training on the real dataset $\mathcal{D}$. According to the additivity of influence functions \cite{group_rffect_analysis,group_rffect_improve,OPT}, as shown by the identity $\Big\|\mathcal{I}_{-\mathcal{D}} + \mathcal{I}_{+\mathcal{S}}\Big\| = \Big\|\Big(\theta^*_\mathcal{D} + \mathcal{I}_{-\mathcal{D}} + \mathcal{I}_{+\mathcal{S}} \Big) - \theta^*_\mathcal{D} \Big\|$, minimizing this term is equivalent to minimizing the displacement between the model parameters $\theta^*_\mathcal{D}$ (trained on $\mathcal{D}$) and the parameters obtained after removing $\mathcal{D}$ and adding $\mathcal{S}$. This directly links the optimization to matching the final model outcomes, ensuring a robust, outcome-oriented data synthesis process.
\end{remark}

\vspace{0.1in}\noindent\textbf{Objective Formulation of Inf-Match} By substituting the estimator in Eq. \eqref{eq:proposition addition} and Eq. \eqref{eq:proposition removal} into the objective formulation defined in Eq. \eqref{eq: new object}, we obtain a differentiable objective for dataset distillation, 
\begin{align}  
\mathcal{S}^*&=\arg\min_{\mathcal{S}} \mathcal{J}(\mathcal{S}) \nonumber\\
\text{s.t.}~~&\mathcal{J}(\mathcal{S}) = \Big\| - \sum_{t} \frac{2\eta_t  \sum_{k \geq t} \eta_k}{|\mathcal{D}|} \Big(H_{\mathcal{D}}^t G_{\mathcal{D}}^t \Big) \nonumber\\
&+ \sum_{t} \frac{\eta_t  \sum_{k \geq t} \eta_k}{|\mathcal{D}|} \Big(H_{\mathcal{D}}^t G_{\mathcal{S}}^t +  H_{\mathcal{S}}^t G_{\mathcal{D}}^t\Big) \Big\|, 
\label{eq: new problem}
\end{align} 
where $H^t_{\mathcal{D}}$ and $H^t_{\mathcal{S}}$ are the hessian of the parameter at the $t$-th (real-set) training iteration on the real and the synthetic set respectively, specifically, $H^t_{\mathcal{D}} = \nabla^2_\theta \mathcal{L}(\mathcal{D}, \theta^t_\mathcal{D})$ and $H^t_{\mathcal{S}} = \nabla^2_\theta \mathcal{L}(\mathcal{S}, \theta^t_\mathcal{D})$. 
And $G^t_{\mathcal{D}}$ and $G^t_{\mathcal{S}}$ are the gradient of the parameter at the $t$-th (real-set) training iteration on the real and the synthetic set respectively, specifically, $G^t_{\mathcal{D}} = \nabla_\theta \mathcal{L}(\mathcal{D}, \theta^t_\mathcal{D})$ and $G^t_{\mathcal{S}} = \nabla_\theta \mathcal{L}(\mathcal{S}, \theta^t_\mathcal{D})$. As for the efficient calculation for the Hessian-gradient production, please refer to Eq. \eqref{eq: hessian}.

During distillation, we do not compute gradients or Hessians over the entire dataset.
Instead, we randomly sample mini-batches $B_{\mathcal{D}} \subset \mathcal{D}$ and $B_{\mathcal{S}} \subset \mathcal{S}$ to estimate these quantities, which greatly improves computational efficiency while preserving unbiased gradient estimates.

\vspace{0.1in} \noindent \textbf{Inf-Match Algorithm} 
\label{sec: Implementation Details} 
The overall pipeline of \textbf{Inf-Match} is outlined in Alg.~\ref{algo: Pipeline}.
We first train a base model on the real dataset $\mathcal{D}$ for $T$ iterations and record the checkpoints ${({\theta}^t_{\mathcal{D}}, \eta_t)}{t=1}^T$.
The synthetic set $\mathcal{S}$ is initialized with real images from $\mathcal{D}$ following the IPC (images-per-class) setting.
At each iteration, we update both the synthetic images and labels by minimizing $\mathcal{J}(\mathcal{S})$ in Eq.~\eqref{eq: new problem} via gradient descent.
To reduce memory consumption, we compute the loss using random mini-batches $B_\mathcal{S} \subset \mathcal{S}$  and $B_\mathcal{D} \subset \mathcal{D}$ from the synthetic and real datasets, respectively.
After convergence, we output the learned synthetic set $\mathcal{S}$.
Below, we detail the initialization and time-step sampling strategies used during training.

\begin{figure*}[tp] 
\centering  
\includegraphics[width=0.99990998199936\linewidth]{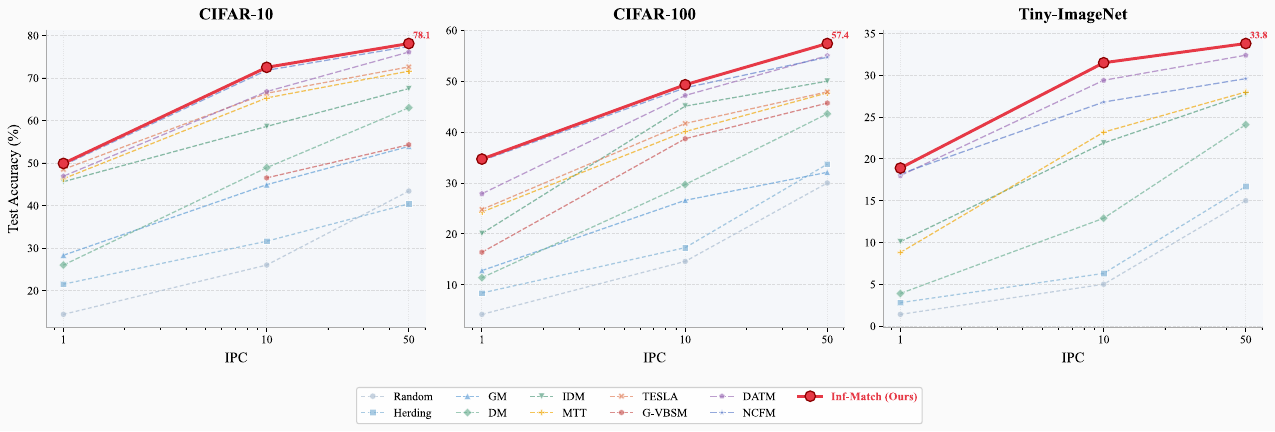} 
\caption{\label{tb:cls} Performance comparison of different methods on CIFAR-10, CIFAR-100, and Tiny-ImageNet. 
} 
\end{figure*}

First, to initialize $\mathcal{S}$, we sample real images from $\mathcal{D}$ according to the IPC (Image Per Class) configuration and assign each a soft label produced by the final model ${\theta}^T_{\mathcal{D}}$.
Both the synthetic images and labels are treated as learnable variables during training, following common practices in dataset distillation~\cite{wang2018dataset,bohdal2020flexible,guo2023towards}.
Compared with one-hot labels, soft labels allow inter-class information sharing, improving the representation efficiency of the distilled data.  

Second, the loss in Eq.~\eqref{eq: new problem} involves averaging over all $T$ training steps.
To improve efficiency, we approximate this average by sampling $m$ checkpoints at each update.
We adopt a time-step selection schedule similar to DATM~\cite{guo2023towards}: early in training, we sample earlier checkpoints to capture basic patterns, while in later stages, we shift toward later checkpoints to encode more complex, fine-grained structures.
This progressive sampling strategy encourages structured learning from simple to difficult information, improving both convergence and final performance.

\section{Experiments} 

This section presents the experimental evaluation of our {Inf-Match} on two scenarios: image classification and vision-language datasets. We first describe the experimental setups, including baselines and implementation details. We then report and analyze the experimental results on several datasets. Finally, we conduct an ablation study to investigate the effect of key components of our method.

\subsection{Experimental Settings}

\vspace{0.2cm}
\noindent\textbf{Baselines.} Here, we choose several baselines as competitors to our approach, including DD \cite{wang2018dataset},  GM \cite{zhaogm}, MTT  \cite{cazenavette2022dataset}, DM \cite{zhaodm}, IDM \cite{zhao2023improved}, DATM \cite{guo2023towards}, TESLA \cite{cui2023scaling}, G-VBSM \cite{shao2023generalized}, NCFM \cite{wang2025datasetdistillationneuralcharacteristic}, and also the famous coreset selection method Herding \cite{Herding}, and the popular baseline random selection.

\vspace{0.2cm}
\noindent\textbf{Network Architecture.}
Our experiments default to the ConvNet architecture, consisting of three (four for Tiny ImageNet) conv-blocks (128 filters, pooling, ReLU, normalization) and a linear classifier. We also investigated LeNet \cite{lenet}, AlexNet \cite{AlexNet}, VGG11 \cite{VGG}, and ResNet18 \cite{resnet} for cross-architecture comparisons. For vision-language tasks, we utilized a pre-trained, trainable Vision Transformer \cite{vit} and a frozen BERT \cite{devlin2018bert}  as backbones; both were independently pre-trained on unimodal data and connected via a trainable linear projection layer.

\vspace{0.2cm}
\noindent\textbf{Setups.} 
We evaluate our method under three different settings of Images Per Class (IPC): 1, 10, and 50. For vision-language datasets, we assess our approach across three synthetic dataset sizes: 100, 500, and 1000 samples. Each experiment consists of two distinct phases. First, we learn from the synthetic data and subsequently train a model on it to evaluate its performance on the real test set. 
All experiments are conducted using the PyTorch framework on a computing server with 8*A100 GPUs. Throughout all experiments, we set the batch size to 50. The optimizer employed in our experiments is SGD-M, with a momentum parameter configured at 0.9. For synthetic images, we set the corresponding learning rate as a value of 50.0, while we set the learning rate for the soft labels as 7.0. To ensure the robustness of our findings and the fairness of the evaluation, each experiment is independently repeated 10 times.

\subsection{Image Classification}

We conduct experiments for image classification datasets like CIFAR-10, CIFAR-100, and Tiny-ImageNet. 
The CIFAR-10 dataset \cite{CIFAR} consists of 60,000 images across 10 different classes, while the CIFAR-100 dataset \cite{CIFAR} is an extension of CIFAR-10, also containing 60,000 32×32-pixel color images divided into 100 fine-grained classes. The Tiny-ImageNet \cite{tiny} dataset, containing 100000 training images, is a subset of the ImageNet dataset. It provides a more challenging dataset than the CIFAR series regarding the number of classes and the complexity of the images.

\subsubsection{Main results}

Figure \ref{tb:cls} presents the impressive performance of our proposed method ("Ours") compared to various dataset distillation techniques across the CIFAR-10, CIFAR-100, and Tiny-ImageNet benchmarks. Our Inf-Match approach consistently achieves the leading results across every configuration of Images Per Class (IPC), establishing the effectiveness and robustness of our outcome-alignment strategy. 
For \textbf{CIFAR-10}, our method achieves the highest accuracy in all settings. It reaches $\mathbf{72.5\%}$ at IPC=10 and $\mathbf{78.1\%}$ at IPC=50, securing an advantage of approximately 0.7\% over the previously top-performing NCFM \cite{wang2025datasetdistillationneuralcharacteristic}. Our $\mathbf{49.9\%}$ accuracy at IPC=1 also represents a new high watermark for extremely low IPC settings. Similarly, on \textbf{CIFAR-100}, our method maintains its dominance, leading all competitors with $\mathbf{49.3\%}$ at IPC=10 and a significant $\mathbf{57.4\%}$ at IPC=50. This superior performance is particularly notable at IPC=50, where we surpass NCFM's result by a margin of $\mathbf{2.7\%}$. The most pronounced gains are observed on \textbf{Tiny-ImageNet}, where our method delivers consistently superior performance across the board. At IPC=10 and IPC=50, our method achieves $\mathbf{31.5\%}$ and $\mathbf{33.8\%}$ respectively, outpacing NCFM by substantial margins of approximately \textbf{4.7\%} and \textbf{4.2\%}. Collectively, these results underscore the power of directly matching model influence, demonstrating our method’s ability to deliver the best performance across various datasets and compression rates.

\subsubsection{Generalization Evaluation}

\begin{figure}[tp] 
\centering  
\includegraphics[width=0.999998199936\linewidth]{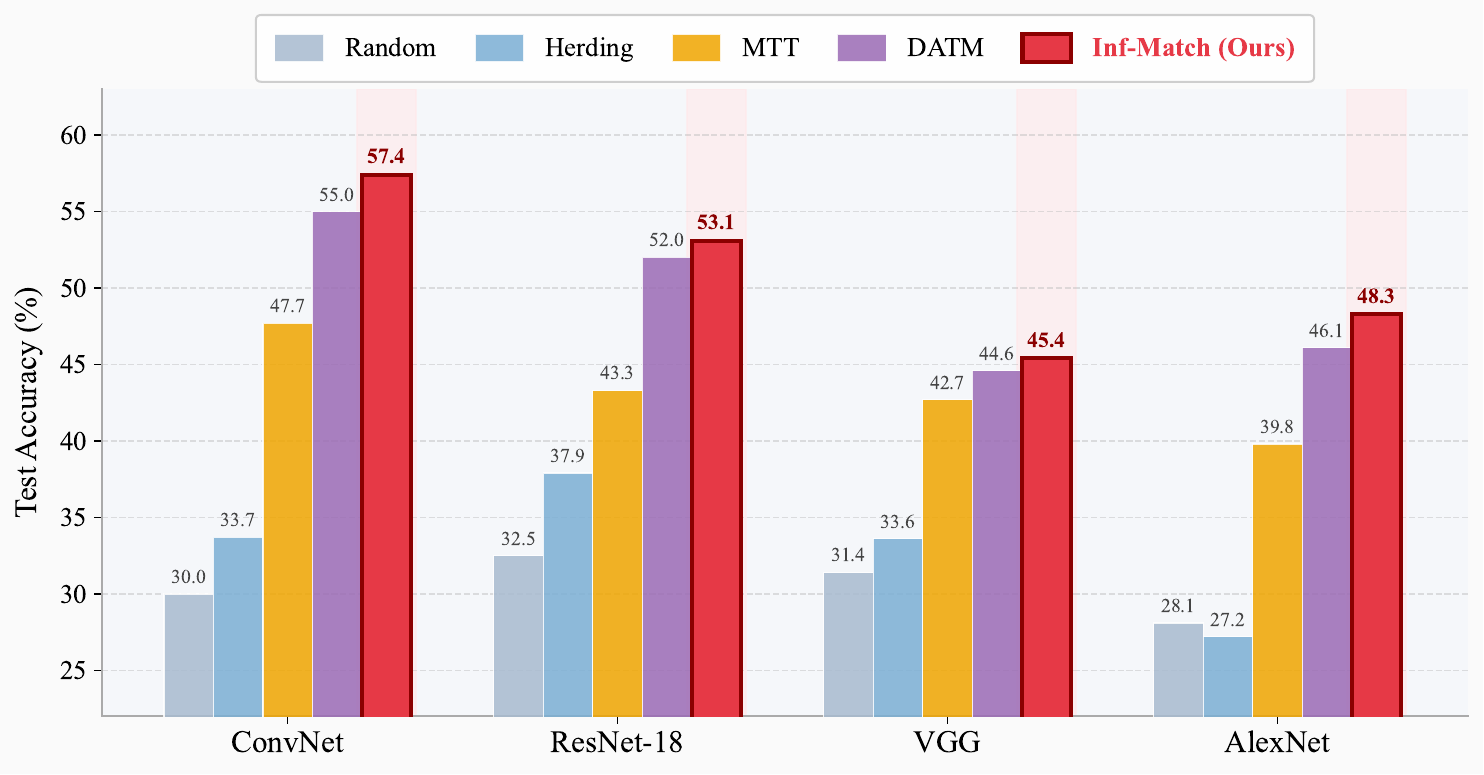} 
\caption{\label{tb:gene}The cross-architecture generalization test on CIFAR-100 with the setting of IPC=50.
} 
\end{figure}

We subsequently assessed the cross-network generalization capability of the synthetic data generated by our method on the CIFAR-100 dataset using an IPC setting of 50. See Table \ref{tb:gene} for details. Importantly, we observe that all distillation schemes presented in the table significantly surpass the performance of selection-based methods, such as random selection and Herding selection \cite{Herding}, in terms of cross-structure generalization. When compared with distillation-based methods (MTT \cite{cazenavette2022dataset} and DATM \cite{guo2023towards}). 
Across various model architecture configurations, our approach consistently outperformed the previous best method, DATM \cite{guo2023towards}, achieving scores between 45.4\% and 57.4\%. This notable improvement underscores the efficacy of our method in producing synthetic data that generalizes effectively across diverse network architectures.

% \begin{table}[tp]
%     \centering
%     \caption{\label{tb:gene}The cross-architecture generalization test on CIFAR-100 with the setting of IPC=50.}  
%     \resizebox{0.8195\linewidth}{!}{
%     \begin{tabular}{l|ccccccc}
%     \toprule
%     Method &ConvNet &ResNet18 &VGG &AlexNet \\
%     \midrule
%     Random &30.0 &32.5 &31.4 &28.1 \\
%     Herding \cite{Herding} &33.7 &37.9 &33.6 &27.2 \\
%     \midrule
%     MTT \cite{cazenavette2022dataset} &47.7 &43.3 &42.7 &39.8 \\
%     DATM \cite{guo2023towards} &55.0 &52.0 &44.6 &46.1 \\
%     \midrule
%     \rowcolor{exp_table_blue}Inf-Match (ours) &\textbf{57.4} &\textbf{53.1} &\textbf{45.4} &\textbf{48.3} \\
%     \bottomrule
%     \end{tabular}
%     } 
% \end{table}

\subsection{Vision-language Datasets}

The field of multimodality \cite{zhao2025equipping,liu2023mars3d,chen2025aligning,10656568,xia2025dreamomni2multimodalinstructionbasedediting,Wu_2025_ICCV} has also made significant progress recently. 
We also conduct experiments for the well-known vision-language dataset Flickr30K \cite{flickr30k}, which comprises 31,783 images depicting daily activities and scenes sourced from the website. Each image is paired with five textual descriptions. 
We follow the experimental settings in BTM \cite{wu2023multimodal}, where we choose the Normalizer-free ResNet \cite{NFNet} as the vision encoder and the BERT \cite{devlin2018bert} model as the text encoder. While both encoders are pretrained, they are trained only on unimodal data and have no exposure to the other modality. A trainable linear projection layer with random initialization follows each encoder. The training loss for the model and the vision language dataset encourages the similarities between those paired images and texts \cite{wu2023multimodal}. 
We selected BTM \cite{wu2023multimodal} as the baseline. BTM introduces a bi-trajectory matching loss for dataset distillation in vision-language datasets, building upon the well-known trajectory matching technique (MTT \cite{cazenavette2022dataset}). As a result, we can readily adapt the updated version of the MTT algorithm, for example, DATM (difficulty-aware trajectory matching) \cite{guo2023towards}, to BTM. Additionally, we have chosen several other baselines, including random selection, Herding selection \cite{Herding}, and GM \cite{zhaogm}. 

\begin{figure}[tp] 
\centering  
\includegraphics[width=0.999998199936\linewidth]{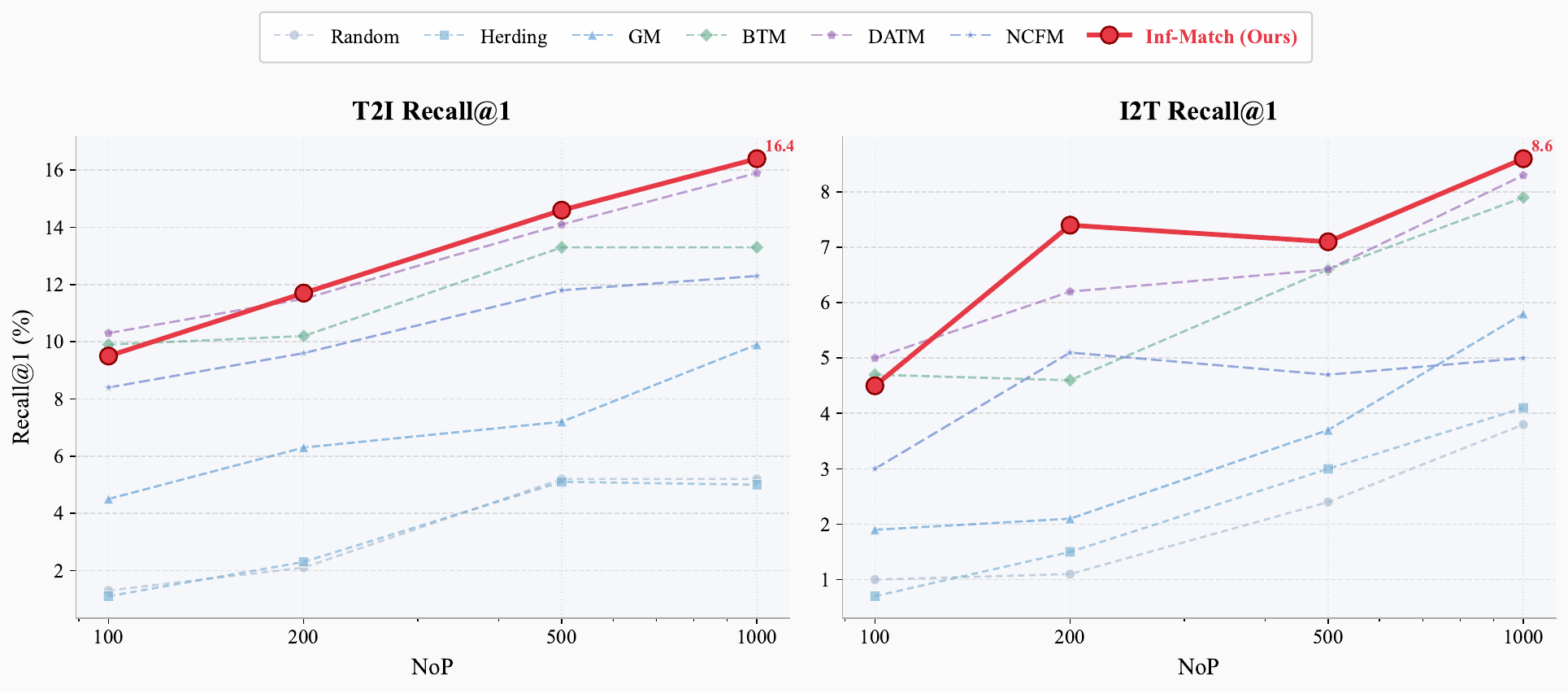} 
\caption{\label{tb:vl} Performance comparison of different dataset distillation methods on the vision-language dataset, Flickr30K \cite{flickr30k}. NoP means the number of (image-text) pairs. T2I and I2T indicate text-to-image and image-to-text, respectively. 
} 
\end{figure}

We provide the experimental results in Figure \ref{tb:vl}. 
In most settings, our method consistently outperformed the others across all configurations. 
For instance, with 200 samples, our method achieved an impressive score of 7.4\% on image-to-text retrieval tasks, higher than the next best, DATM \cite{guo2023towards}, at 1.3\%. 
With 500 samples, our method recorded 14.6\% on the text-to-image setting, leading the pack, while the second-best competitor, DATM, managed only 14.1\%. 
Finally, at 1000 samples, we maintained our superiority with a score of 16.4\% on the text-to-image setting, surpassing all other methods. 
Another noteworthy observation from the experiments is that all dataset distillation methods outperformed the coreset methods based on selection. This finding aligns with previous results in Figure \ref{tb:cls} and strongly indicates that dataset distillation is a promising direction for further research. 
Overall, our approach demonstrates robust performance across various sample sizes, highlighting its effectiveness on vision-language datasets.

\begin{table}[tp]
    \centering
    \caption{Ablation study on CIFAR-100 with IPC=50. We also select NCFM \cite{wang2025datasetdistillationneuralcharacteristic}, DATM\cite{guo2023towards}, DM \cite{zhaodm}, GM \cite{zhaogm}, MTT \cite{cazenavette2022dataset} as baselines.  
    }  
    \resizebox{0.999\linewidth}{!}{
    \begin{tabular}{ccc|cccc}
    \toprule
    Real-data Init. &Learnable-Label &Sampling-Schedule &Performance \\
    \midrule
    \rowcolor{exp_table_blue}$\times$ & $\times$ &$\times$   &52.2\\
    \midrule
    \multicolumn{3}{c|}{GM \cite{zhaogm}} &32.1\\
    \multicolumn{3}{c|}{DM \cite{zhaodm}} &43.6\\
    \multicolumn{3}{c|}{MTT \cite{cazenavette2022dataset}} &47.7\\
    \midrule
    \rowcolor{exp_table_blue}$\checkmark$ & $\times$ &$\times$   &53.7\\
    \rowcolor{exp_table_blue}$\checkmark$ &$\times$  & $\checkmark$  &55.0\\
    \rowcolor{exp_table_blue}$\checkmark$ & $\checkmark$ &$\times$   &54.6\\
    \rowcolor{exp_table_blue}$\checkmark$ & $\checkmark$ &$\checkmark$  &\textbf{57.4}\\
    \midrule
    \multicolumn{3}{c|}{DATM \cite{guo2023towards}} &55.0\\
    \multicolumn{3}{c|}{NCFM \cite{wang2025datasetdistillationneuralcharacteristic}} &54.7\\
    \bottomrule
    \end{tabular}
    }
    \label{tab:ablation} 
\end{table}

\subsection{Ablation study}  

Table \ref{tab:ablation} details the ablation study of our Inf-Match framework on CIFAR-100 ($\text{IPC}=50$), confirming the contribution of each component while validating the strength of our core methodology.  
The sequential incorporation of enhancements improves performance: introducing Real-data Initialization boosts accuracy to $53.7\%$, utilizing the Sampling Schedule achieves $55.0\%$, and including Learnable Labels reaches $54.6\%$. When all components are integrated, our method attains its final accuracy of $\mathbf{57.4\%}$. This not only confirms the positive synergistic effect of our components but also robustly outperforms other advanced distillation methods like DATM ($\mathbf{55.0\%}$) and NCFM ($\mathbf{54.7\%}$), underscoring that our outcome-alignment strategy yields a superior solution space compared to process-matching techniques.

\subsection{Learning process visualization}

\begin{figure}[tp]
\centering 
%\hspace{-1.025cm}
\includegraphics[width=0.9588\linewidth]{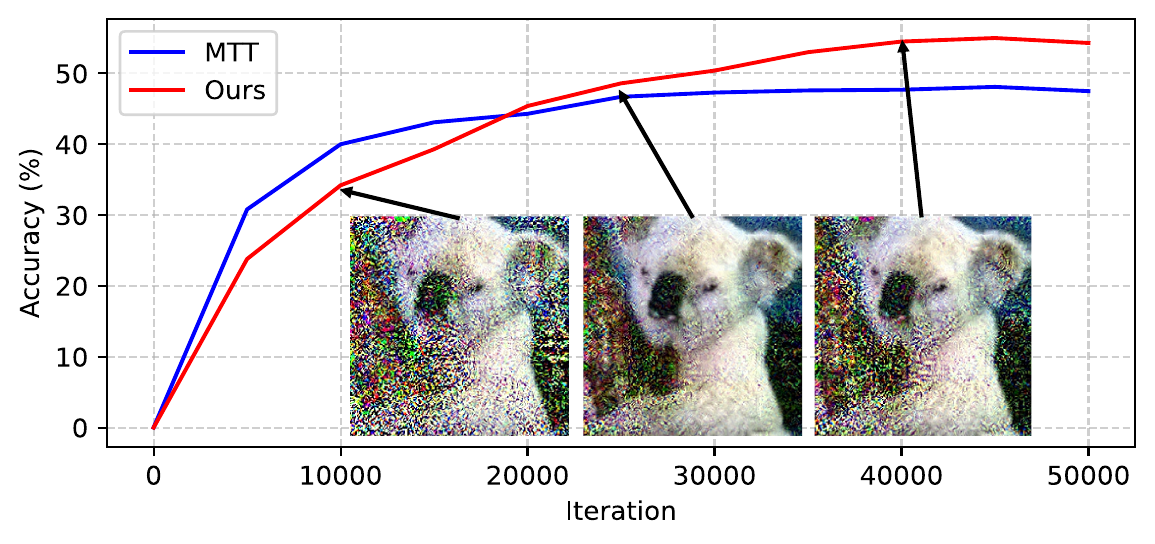} 
\caption{\label{fig: ablation 2}
Visualization of the learning curve (MTT \cite{cazenavette2022dataset} v.s. Ours) and the intermediate synthetic results of ours. This experiment is conducted on CIFAR-100 with the setting of IPC=50. 
} 
\end{figure}

In Figure \ref{fig: ablation 2}, we also conducted a comparative visualization of the learning process from synthetic data during algorithm execution. Firstly, we compare the performance as iterations progress. It is evident that, compared to the heuristic-based proxy task, \textit{e.g.} MTT \cite{cazenavette2022dataset}, our approach, which directly optimizes the original problem, exhibits a slower convergence rate but ultimately achieves significantly better performance. Secondly, we found that the fidelity of the synthetic images is not necessarily correlated with the performance of the final synthetic data. This observation arises from our findings that during the learning process, the images synthesized by our approach undergo a transition from increased realism to a subsequent increase in noise.

\subsection{Feature space visualization}

\begin{figure}[tp]
\centering  
\includegraphics[width=1.032\linewidth]{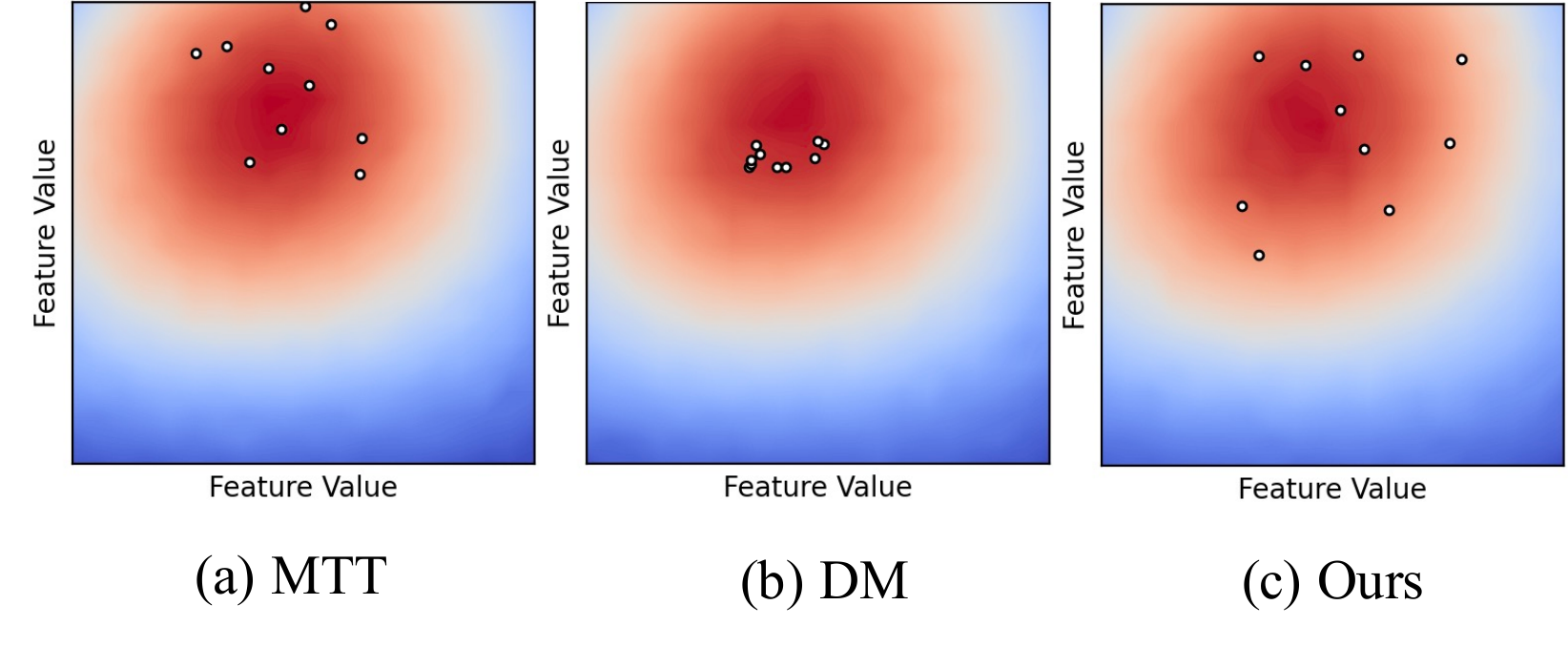} 
\caption{\label{fig: comp featue} 
An illustration of why our method performs well for dataset distillation is provided by comparing the feature distribution of our method and the classic DM (distribution matching) \cite{zhaodm} and MTT (matching training trajectory) \cite{cazenavette2022dataset}.  
} 
\end{figure}

Furthermore, we compare the learned condensed dataset in the embedding space to study their distributions, see Fig. \ref{fig: comp featue}. We visualized the feature space of the ``\textit{Wolf}" category in CIFAR-100 under the IPC=10 setting. The white scatter points represent the projections of synthetic samples within this feature space. We found that when using DM \cite{zhaodm} with the goal of feature distribution matching at small IPC settings (IPC=10), the synthetic samples produced tend to be overly concentrated in the high-density regions of the real data. In contrast, our approach generates synthetic samples that effectively occupy both high-density areas and the less dense regions at the distribution's edges, demonstrating a more balanced representation compared to MTT and DM.

\section{Conclusion}

We introduced a novel dataset distillation framework that redefines the optimization objective from heuristic process alignment to rigorous outcome alignment. By developing an efficient, accurate, and fully differentiable sample influence estimator, we were able to precisely quantify the contribution of both the real and synthetic datasets to the final model parameters. Extensive experiments demonstrated the superior performance over state-of-the-art distillation methods.

\section{Acknowledgment}

The work has been supported by Hong Kong Research Grant Council-General Research Fund Scheme (Grant No. 17202422, 17212923, 17215025), Theme-based Research (Grant No. T45-701/22-R), and Strategic Topics Grant (Grant No. STG3/E-605/25-N). Part of the described research work is conducted in the JC STEM Lab of Robotics for Soft Materials funded by The Hong Kong Jockey Club Charities Trust.

{
    \small
    \bibliographystyle{ieeenat_fullname}
    \bibliography{main}
}

\end{document}